\documentclass{article}
\usepackage{graphicx} 
\usepackage{booktabs}  
\usepackage[backend=bibtex,style=numeric,sorting=none]{biblatex}
\usepackage{hyperref}
\usepackage{multirow}
\usepackage{amsmath}
\usepackage{longtable}  
\usepackage{array} 
\usepackage{changepage} 
\usepackage[margin=1in]{geometry}
\usepackage{todonotes}
\usepackage{threeparttable}
\hypersetup{
    colorlinks=true,        
    linkcolor=black,        
    citecolor=blue,         
    urlcolor=blue,          
    pdfborder={0 0 0},      
    hyperfootnotes=false    
}
\usepackage{adjustbox,soul}

\title{AI Diffusion in Low Resource Language Countries}
\author{Amit Misra, Syed Waqas Zamir, Wassim Hamidouche, \\ Inbal Becker-Reshef and Juan Lavista Ferres \\ \\ Microsoft AI for Good Research Lab}
\date{}

\begin{document}

\maketitle

\begin{abstract}
Artificial intelligence (AI) is diffusing globally at unprecedented speed, but adoption remains uneven. Frontier Large Language Models (LLMs) are known to perform poorly on low-resource languages due to data scarcity. We hypothesize that this performance deficit reduces the utility of AI, thereby slowing adoption in Low-Resource Language Countries (LRLCs). To test this, we use a weighted regression model to isolate the language effect from socioeconomic and demographic factors, finding that LRLCs have a share of AI users that is approximately 20\% lower relative to their baseline. These results indicate that linguistic accessibility is a significant, independent barrier to equitable AI diffusion.
\end{abstract}

\section{Introduction}

Artificial intelligence is the latest general‑purpose technology and is diffusing across the globe at historic speed~\supercite{bick2024rapid}. 
However, as with past waves of technology like electricity and the internet, adoption is not uniform across countries. Familiar patterns of digital inequality are emerging: countries with stronger economies, higher education levels, and better digital infrastructure showing higher and earlier AI adoption~\supercite{khan_artificial_2024, liu_who_2024}.

A key difference with AI, however, is that language is not merely an interface, it is the substrate of the technology’s functionality. Modern AI, especially large language models (LLMs), relies on learning from vast amounts of text data. The open web, a primary source for training LLMs, is heavily skewed toward a few high-resource languages. Nearly half of all web content is in English~\supercite{common_crawl_lang_data}, only 5.1\% of the global population are native English speakers~\supercite{cia_world_factbook}. Other major languages (such as Spanish, French, German, Russian, Mandarin Chinese) have roughly 10× less web data than English, yet still enjoy a strong digital presence~\supercite{common_crawl_lang_data,penedo2025fineweb2pipelinescale}. Among these, languages in the Indo-European family, in particular, may also benefit from cultural and linguistic similarities with English. In contrast, the overwhelming majority of the world’s $\sim$7,000 languages are low-resource, with minimal to no digital footprint~\supercite{penedo2025fineweb2pipelinescale}. The imbalance in data directly translates into LLM performance disparities. Multilingual benchmark tests consistently find that LLMs perform poorly on low-resource languages (LRLs) relative to high-resource ones\supercite{buscemi2025mindlanguagegapautomated, singh2025globalmmluunderstandingaddressing, adelani2025irokobenchnewbenchmarkafrican}.

In this paper, we present what is, to our knowledge, the first country‑level assessment of how language resourcing relates to AI diffusion. We classify countries by the prevalence of low‑resource languages—defining (LRLCs) via a tiered scheme that captures language resourcing and dominant language(s) in a country—and merge this classification with usage telemetry from 147 countries/economics~\supercite{misra_wang_ai_diffusion}, while controlling for socioeconomic and demographic covariates (GDP per capita, electricity access, internet penetration, and age structure). Using cross‑country regressions and per‑capita usage metrics, we compare AI adoption in LRLCs versus higher‑resource‑language countries.

Our results reveal a sizable language-linked gap: in raw terms, LRLCs exhibit less than half the per-capita use of AI tools. After controlling for covariates, we estimate an $\sim$20\% lower share of AI users is attributable to language factors. We also analyze recent trends to assess whether this gap is narrowing as systems improve. Taken together, the findings indicate that while income and digital infrastructure are necessary, they are not sufficient for inclusive AI diffusion; linguistic accessibility must be addressed. We conclude with implications for research and policy to promote AI adoption with—rather than at the expense of—marginalized languages. 

\section{Methods}
\label{sec:methods}
\subsection{Defining Low Resource Language Countries}
\subsubsection{Language Classification}
To quantify AI diffusion readiness by country, we first construct a taxonomy of languages based on the availability of digital content and performance in frontier LLMs. We leverage the FineWeb2 multilingual dataset~\supercite{penedo2025fineweb2pipelinescale}, which contains text corpora covering over 1,000+ languages, to categorize each language into one of three groups:
\begin{itemize}
    \item \textbf{High-resource languages:} This category comprises languages with large-scale digital presence and strong representation in existing LLMs. Examples include English, Spanish, French, German, Italian, Russian, Portuguese, Mandarin Chinese and Japanese.
    \item \textbf{Mid-resource languages:} These are languages with moderate amounts of text data available online. LLMs can perform reasonably in these languages, however not as accurately as in high-resource languages. Examples include Arabic, Hindi, Bengali, Polish, Dutch, Indonesian, Vietnamese, Persian, Turkish, Thai, Korean, Ukrainian, Greek, Czech, Swedish, Hungarian, Danish, Finnish, Hebrew, Malay, and many others.
    \item \textbf{Low-resource languages:} This category encompasses the majority of the world's languages, those with very limited textual data, or none at all. Examples include Chichewa, Inuktitut and Guarani. 
\end{itemize}

\subsubsection{Country-Level Classification}
With the above language taxonomy defined, we assign each country an “AI diffusion readiness” label (High, Intermediate, Low) based on its dominant national language. Our primary reference for country languages is the CIA World Factbook~\supercite{cia_factbook}, which provides (regularly updated) information on official and widely spoken languages in each country. For each country, we apply the following steps. 

\vspace{0.2em}
\noindent\textbf{(1) Language extraction:} We retrieve the languages field from the World Factbook. Since this language field contains unstructured data, we use GPT-5 reasoning model as parser to extract the dominant language name for that country. 

\vspace{0.2em}
\noindent\textbf{(2) Map the dominant language to a resource category:} We check which of our three categories  the identified language fell into. If the language is in our predefined high resource list, the country is labeled as a High-Resource Language Country (HRLC). If the language is in the Intermediate list, the country receives  Mid-Resource Language Country (MRLC) label. All other countries default to Low-Resource Language Country (LRLC) category.

\subsection{Estimating the Impact of Low-Resource Languages on AI Diffusion}

To estimate the impact of LRLC status on AI adoption, we model the relationship between the two using AI User Share, a Microsoft metric estimating the proportion of the working-age population using AI per country~\supercite{misra_wang_ai_diffusion}. Our primary estimate comes from a fractional logit Generalized Linear Model (GLM), weighted to produce the Average Treatment effect on the Treated (ATT)~\supercite{papkeWooldridge1996}. This model is well-suited for a bounded outcome (a percentage) and estimates the change in AI usage if an LRLC were to become a non-LRLC. In all models, we control for GDP per capita (log scale)~\supercite{world_bank_open_data}, electricity access~\supercite{Worldbank_electricity}, internet access~\supercite{itu_internet}, and age structure~\supercite{cia_factbook}. Continuous covariates are standardized before propensity score estimation to stabilize optimization.

To ensure our findings are robust, we compare the GLM results against several alternative estimators, including Inverse Probability Weighting (IPW), Augmented IPW (AIPW), and a standard Ordinary Least Squares (OLS) regression~\supercite{hiranoImbensRidder2000, glynnQuinn2009, kurzAIPW2021}. As shown in Table \ref{tab:methods_comparison}, these methods consistently point to a meaningful negative impact for LRLCs and are consistent with the ATT-weighted GLM results.

\begin{table}[htbp]
\centering
\begin{threeparttable}
\caption{Comparison of Treatment Effect Estimates for Low-Resource Language Status}
\label{tab:methods_comparison}
\begin{tabular}{lcccc}
\toprule
Method & Estimate & Std.\ Error & 95\% CI & ESS \\
\midrule
OLS                & $-1.88$ & $0.96$ & $[-3.77,\ \phantom{-} 0.01]$ & -- \\
ATT (IPW)          & $-2.32$ & $0.80$ & $[-3.89,\ -0.75]$ & 21.1 \\
ATT (AIPW)         & $-1.69$ & $0.91$ & $[-3.48,\ \phantom{-} 0.09]$  & 21.1 \\
ATT-weighted GLM   & $-2.07$ & $0.86$ & $[-3.76,\ -0.38]$ & 21.1 \\
\bottomrule
\end{tabular}
\begin{tablenotes}\footnotesize
\item Effects are in percentage points on AI User Share. ESS = Kish effective sample size for controls. CIs use robust (HC3) SEs for OLS and ATT‑weighted GLM and percentile bootstrap for ATT estimators. 
\end{tablenotes}
\end{threeparttable}
\end{table}

For weighting methods, propensity scores were fit via L2‑regularized logistic regression and clipped to [0.02,0.98] to avoid extreme weights. Uncertainty is quantified via stratified bootstrap at the country level (1,000 draws) with 95\% percentile Confidence Intervals (CI)s~\supercite{efron1979}. Impact estimates are reported in percentage points (pp), but we also present the main effect relative to the LRLC mean for easier interpretation.

To assess whether gaps widened or narrowed, we estimate a two‑period difference‑in‑differences (2024 vs.\ 2025) ATT for LRLCs. For the panel, we apply an AIPW‑DiD estimator~\supercite{SantAnnaZhao2020} that combines pre‑period covariate weighting with outcome regression for double robustness, using the same covariates  as above. Diagnostics include propensity score overlap, standardized mean differences (before/after ATT weighting)~\supercite{austin2011ps} with a nominal $|\mathrm{SMD}|<0.1$ threshold and the control effective sample size (ESS)~\supercite{kish1965}. As a robustness check, we also estimate IPW‑DiD; results are directionally consistent with AIPW‑DiD. 

\section{Results}
Table~\ref{tab:ai_usage_raw} presents the raw, unmodeled differences between LRLCs and non-LRLCs in AI User Share for the earliest and latest available data points, along with the relative change. The AI User Share for LRLCs has consistently been less than half that of other countries. While both groups have increased over time, the growth for LRLCs is smaller in both absolute and relative terms.

\begin{table}[htbp]
\centering
\begin{threeparttable}
\caption{AI Usage Share and Relative Change by Country Type}
\label{tab:ai_usage_raw}
\begin{tabular}{lccc}
\toprule
\textbf{Language Category} & \textbf{AI User Share (2025)} & \textbf{AI User Share (2024)} & \textbf{Relative Change} \\
\midrule
non-LRLCs  & 21.3\% & 17.2\% & 23\% \\
LRLCs & \phantom{0}9.9\%  & \phantom{0}8.5\%  & 17\% \\
\bottomrule
\end{tabular}
\begin{tablenotes}[flushleft]\footnotesize
\item AI User Share is the percentage of users using AI tools in each country type.  
\item Relative Change is the percentage increase from 2024 to 2025 within each group.
\end{tablenotes}
\end{threeparttable}
\end{table}

The raw data in Table~\ref{tab:ai_usage_raw}, however, are confounded by factors like GDP per capita. To isolate the effect of language, we turn to our model-based estimates. After adjusting for socioeconomic and demographic covariates, our primary model reveals a significant adoption gap for LRLCs, as detailed in Table~\ref{tab:ai_adoption_levels}. For 2025, we estimate a 2.1 percentage point (pp) lower share of AI users in LRLCs. Relative to the LRLC baseline of 10 pp, this represents a substantial $\sim$20\% shortfall in adoption. This estimate appears to be conservative; sensitivity analyses using alternative country classification rules consistently yielded similar or larger effects. This analysis confirms that the gap between LRLCs and non-LRLCs is not merely a reflection of income or infrastructure, but is independently driven by language factors.
\begin{table}[htbp]
\centering
\caption{AI User Adoption Rates: Low Resource Language Countries vs. Others}
\label{tab:ai_adoption_levels}
\resizebox{\textwidth}{!}{
\begin{tabular}{lcccccccc}
\toprule
\textbf{Year} & \textbf{LRLC Baseline} & \textbf{Gap Estimate} &
\multicolumn{3}{c}{\textbf{Counterfactual}} &
\multicolumn{3}{c}{\textbf{Rel. Impact (\%)}} \\
& (pp) & (pp) & Point Est. & CI Low & CI High & Point Est. & CI Low & CI High \\
\midrule
2025 & 10.0 & 2.1 & 12.1 & 10.4 & 13.8 & 21 & \phantom{-}4 & 38\\
2024 & \phantom{0}8.6  & 1.2 & \phantom{0}9.8  & \phantom{0}8.3  & 11.4 & 14 & -4 & 33\\
\bottomrule
\end{tabular}
}
\begin{flushleft} \footnotesize
LRLC Baseline shows observed AI adoption rates in Low Resource Language Countries in percentage points (pp). Gap Estimate represents the estimated treatment effect (gap, pp). Counterfactual shows the estimated adoption rate LRLCs would achieve without resource constraints (LRLC Baseline + Gap Estimate, pp). 95\% CI gives confidence interval for the counterfactual estimate. Relative Impact shows the treatment effect as a percentage of the LRLC baseline adoption rate, and the corresponding 95\% CI.
\end{flushleft}

\end{table}

We assessed changes in the relative difference between LRLCs and non-LRLCs by estimating the impact using 2024 data and by conducting a Difference-in-Differences (DiD) analysis. As shown in Table~\ref{tab:ai_adoption_levels}, the estimated relative impact for 2024 is smaller than for 2025, although the 95\% confidence intervals substantially overlap. Furthermore, the DiD analysis detects no statistically significant change in the treatment effect over time (0.08 pp; 95\% CI: [–1.21, 2.00] percentage points). This suggests that the apparent widening may reflect common temporal trends affecting both groups rather than a causal expansion of the digital divide. 

\section{Discussion and Conclusions}

Our results show that LRLCs have systematically lower AI usage than non-LRLCs, even after controlling for GDP per capita, electricity, internet access, and population age distribution by country. Since LLMs perform poorly in low-resource languages due to limited training data, it is unsurprising that this performance gap translates into lower adoption.

Furthermore, we find no statistically significant evidence that the usage gap is either widening or narrowing. LRLCs are not catching up to non-LRLCs, raising the risk that countries already lagging on key economic indicators like GDP per capita will fall further behind as they fail to fully capture AI’s benefits.

The only solution is to build high-quality training datasets for low-resource languages. From a technical standpoint, there is no workaround: without sufficient data, LLMs cannot even reach reasonable performance in these languages. Until that gap is closed, millions of people will remain excluded from the full potential of AI in their native language.

Our study has several limitations. While defining low-resource versus non-low-resource languages is relatively straightforward, mapping these categories onto countries is more complex. Many people worldwide are multilingual, and reliable data on second-language speakers by country is often unavailable or incomplete. Moreover, even in countries with high-resource official languages, low literacy rates can limit large segments of the population from using AI. Significant disparities may also exist within the same country, particularly between major cities and rural areas. These factors highlight important directions for future work in refining our country-level classification model. Furthermore, the limited one-year time horizon of our dataset means longer-term data is needed to determine whether the LRLC vs. non-LRLC gap is stable, shrinking, or expanding over time

Overall, our findings underscore a core implication: without deliberate efforts to improve training data for low-resource languages, LRLCs risk being systematically excluded from the benefits of the latest general-purpose technology. Prioritizing high-quality multilingual datasets is essential to ensure inclusive AI diffusion.

\printbibliography  

\clearpage  
\appendix  
\section*{Appendix}  
\addcontentsline{toc}{section}{Appendix}  
\begin{longtable}{@{}ll@{}}
\caption{AI Diffusion Labels by Country} \\
\toprule
\textbf{Country} & \textbf{AI Diffusion Label} \\
\midrule
\endfirsthead
\midrule \multicolumn{2}{r}{Continued on next page} \\
\endfoot
\bottomrule
\endlastfoot
Afghanistan & LRLC \\
Albania & MRLC \\
Algeria & MRLC \\
Andorra & MRLC \\
Angola & HRLC \\
Antigua and Barbuda & HRLC \\
Argentina & HRLC \\
Armenia & LRLC \\
Australia & HRLC \\
Austria & HRLC \\
Azerbaijan & MRLC \\
Bahrain & MRLC \\
Bangladesh & MRLC \\
Barbados & HRLC \\
Belarus & HRLC \\
Belgium & MRLC \\
Belize & HRLC \\
Benin & HRLC \\
Bhutan & LRLC \\
Bolivia & HRLC \\
Bosnia and Herzegovina & MRLC \\
Botswana & LRLC \\
Brazil & HRLC \\
Brunei & MRLC \\
Bulgaria & MRLC \\
Burkina Faso & LRLC \\
Burma & LRLC \\
Burundi & LRLC \\
Cambodia & LRLC \\
Cameroon & HRLC \\
Canada & HRLC \\
Cape Verde & HRLC \\
Central African Republic & LRLC \\
Chad & HRLC \\
Chile & HRLC \\
China & HRLC \\
Colombia & HRLC \\
Comoros & MRLC \\
Congo & HRLC \\
Congo DR & HRLC \\
Costa Rica & HRLC \\
Cote d'Ivoire & LRLC \\
Croatia & MRLC \\
Cuba & HRLC \\
Cyprus & MRLC \\
Czechia & MRLC \\
Denmark & MRLC \\
Djibouti & HRLC \\
Dominica & HRLC \\
Dominican Republic & HRLC \\
Ecuador & HRLC \\
Egypt & MRLC \\
El Salvador & HRLC \\
Equatorial Guinea & HRLC \\
Eritrea & LRLC \\
Estonia & MRLC \\
Eswatini & HRLC \\
Ethiopia & LRLC \\
Fiji & HRLC \\
Finland & MRLC \\
France & HRLC \\
Gabon & HRLC \\
Georgia & MRLC \\
Germany & HRLC \\
Ghana & LRLC \\
Greece & MRLC \\
Grenada & HRLC \\
Guatemala & HRLC \\
Guinea & HRLC \\
Guinea-Bissau & LRLC \\
Guyana & HRLC \\
Haiti & HRLC \\
Honduras & HRLC \\
Hungary & MRLC \\
Iceland & LRLC \\
India & MRLC \\
Indonesia & LRLC \\
Iran & MRLC \\
Iraq & MRLC \\
Ireland & HRLC \\
Israel & MRLC \\
Italy & HRLC \\
Jamaica & HRLC \\
Japan & HRLC \\
Jordan & MRLC \\
Kazakhstan & HRLC \\
Kenya & HRLC \\
Kiribati & LRLC \\
Kosovo & MRLC \\
Kuwait & MRLC \\
Kyrgyzstan & LRLC \\
Laos & LRLC \\
Latvia & MRLC \\
Lebanon & MRLC \\
Lesotho & LRLC \\
Liberia & HRLC \\
Libya & MRLC \\
Liechtenstein & HRLC \\
Lithuania & MRLC \\
Madagascar & LRLC \\
Malawi & LRLC \\
Malaysia & MRLC \\
Maldives & LRLC \\
Mali & HRLC \\
Malta & LRLC \\
Marshall Islands & LRLC \\
Mauritania & MRLC \\
Mauritius & LRLC \\
Mexico & HRLC \\
Micronesia & HRLC \\
Moldova & MRLC \\
Monaco & HRLC \\
Mongolia & LRLC \\
Montenegro & MRLC \\
Morocco & MRLC \\
Mozambique & LRLC \\
Namibia & LRLC \\
Nauru & LRLC \\
Nepal & MRLC \\
Netherlands & MRLC \\
New Zealand & HRLC \\
Nicaragua & HRLC \\
Niger & LRLC \\
Nigeria & HRLC \\
North Korea & MRLC \\
North Macedonia & LRLC \\
Norway & MRLC \\
Oman & MRLC \\
Pakistan & LRLC \\
Palau & LRLC \\
Panama & HRLC \\
Papua New Guinea & LRLC \\
Paraguay & HRLC \\
Peru & HRLC \\
Philippines & LRLC \\
Poland & MRLC \\
Portugal & HRLC \\
Qatar & MRLC \\
Romania & MRLC \\
Russia & HRLC \\
Rwanda & LRLC \\
Saint Kitts and Nevis & HRLC \\
Saint Lucia & HRLC \\
Saint Vincent and the Grenadines & HRLC \\
Samoa & LRLC \\
San Marino & HRLC \\
Sao Tome and Principe & HRLC \\
Saudi Arabia & MRLC \\
Senegal & HRLC \\
Serbia & MRLC \\
Seychelles & LRLC \\
Sierra Leone & LRLC \\
Singapore & HRLC \\
Slovakia & MRLC \\
Slovenia & MRLC \\
Solomon Islands & LRLC \\
Somalia & LRLC \\
South Africa & LRLC \\
South Korea & MRLC \\
South Sudan & HRLC \\
Spain & HRLC \\
Sri Lanka & LRLC \\
Sudan & MRLC \\
Suriname & MRLC \\
Sweden & MRLC \\
Switzerland & HRLC \\
Syria & MRLC \\
Tajikistan & LRLC \\
Tanzania & LRLC \\
Thailand & MRLC \\
The Bahamas & HRLC \\
The Gambia & HRLC \\
Timor-Leste & LRLC \\
Togo & HRLC \\
Tonga & LRLC \\
Trinidad and Tobago & HRLC \\
Tunisia & MRLC \\
Turkey & MRLC \\
Turkmenistan & LRLC \\
Tuvalu & LRLC \\
Uganda & LRLC \\
Ukraine & MRLC \\
United Arab Emirates & MRLC \\
United Kingdom & HRLC \\
United States & HRLC \\
Uruguay & HRLC \\
Uzbekistan & LRLC \\
Vanuatu & LRLC \\
Vatican City (Holy See) & HRLC \\
Venezuela & HRLC \\
Vietnam & MRLC \\
Yemen & MRLC \\
Zambia & LRLC \\
Zimbabwe & LRLC \\
\label{tab:ai_diffusion_labels}
\end{longtable}

\end{document}